\documentclass[10pt,twocolumn,letterpaper]{article}

\usepackage[pagenumbers]{iccv} 

\definecolor{iccvblue}{rgb}{0.21,0.49,0.74}
\usepackage[pagebackref,breaklinks,colorlinks]{hyperref}

\usepackage{times}
\usepackage{epsfig}
\usepackage{graphicx}
\usepackage{amsmath}
\usepackage{amssymb}
\usepackage{subcaption}
\usepackage{booktabs}
\usepackage{multirow}
\usepackage{caption}
\usepackage{float}
\usepackage{makecell}
\usepackage{array}
\usepackage{setspace}
\usepackage{afterpage}
\usepackage{lipsum}
\usepackage{dblfloatfix}
\usepackage{algorithm}
\usepackage{algorithmic}
\usepackage{setspace}
\usepackage{appendix}

\def\paperID{13448} 
\def\confName{ICCV}
\def\confYear{2025}

\title{InternVL-X: Advancing and Accelerating InternVL Series with \\
    Efficient Visual Token Compression}

\author{
    Dongchen Lu\textsuperscript{1}  \hspace{1cm} 
    Yuyao Sun\textsuperscript{2}  \hspace{1cm} 
    Zilu Zhang\textsuperscript{1}  \hspace{1cm} 
    Leping Huang\textsuperscript{1} \\[0.2em]
    Jianliang Zeng\textsuperscript{3} \hspace{1cm} 
    Mao Shu\textsuperscript{1*} \hspace{1cm} 
    Huo Cao\textsuperscript{1*}  \\[0.7em]
    \textsuperscript{1} Baidu Inc. \\
    \textsuperscript{2} Xidian University \\
    \textsuperscript{3} University of Chinese Academy of Sciences \\
    \textsuperscript{*} Corresponding Author
}

\date{
}

\begin{document}
\maketitle

\begin{abstract}

Most multimodal large language models (MLLMs) treat visual tokens as ``a sequence of text”, integrating them with text tokens into a large language model (LLM). However, a great quantity of visual tokens significantly increases the demand for computational resources and time. In this paper, we propose InternVL-X, which outperforms the InternVL model in both performance and efficiency by incorporating three visual token compression methods. First, we propose a novel vision-language projector, PVTC. This component integrates adjacent visual embeddings to form a local query and utilizes the transformed CLS token as a global query, then performs point-to-region cross-attention through these local and global queries to more effectively convert visual features. Second, we present a layer-wise visual token compression module, LVTC, which compresses tokens in the LLM’s shallow layers and then expands them through upsampling and residual connections in the deeper layers. This significantly enhances the model’s computational efficiency. Futhermore, we propose an efficient high resolution slicing method, RVTC, which dynamically adjusts the number of visual tokens based on image area or length filtering. RVTC greatly enhances training efficiency with only a slight reduction in performance. By utilizing 20\% or fewer visual tokens, InternVL-X achieves state-of-the-art performance on 7 public MLLM benchmarks, and improves the average metric by 2.34\% across 12 tasks. The source code is available at \url{https://github.com/ludc506/InternVL-X}.
\end{abstract}

\vspace{-1.5ex}

\section{Introduction}
\vspace{-0.5ex}

MLLMs, such as BLIP \cite{blip,blip2,instructblip}, LLaVA \cite{llava,llava_1_5,llavanext,llavaonevision}, QwenVL \cite{qwenvl,qwen2vl}, InternVL \cite{internvl,internvl_1_5} have demonstrated remarkable progress in recent years, showcasing impressive performance across various vision-language tasks such as visual question answering, image captioning, etc. MLLMs are attracting more attention from the academia and open-source community.


\begin{figure}[t]
\setlength{\belowcaptionskip}{-12pt}
  \centering
  \includegraphics[width=3.25in]{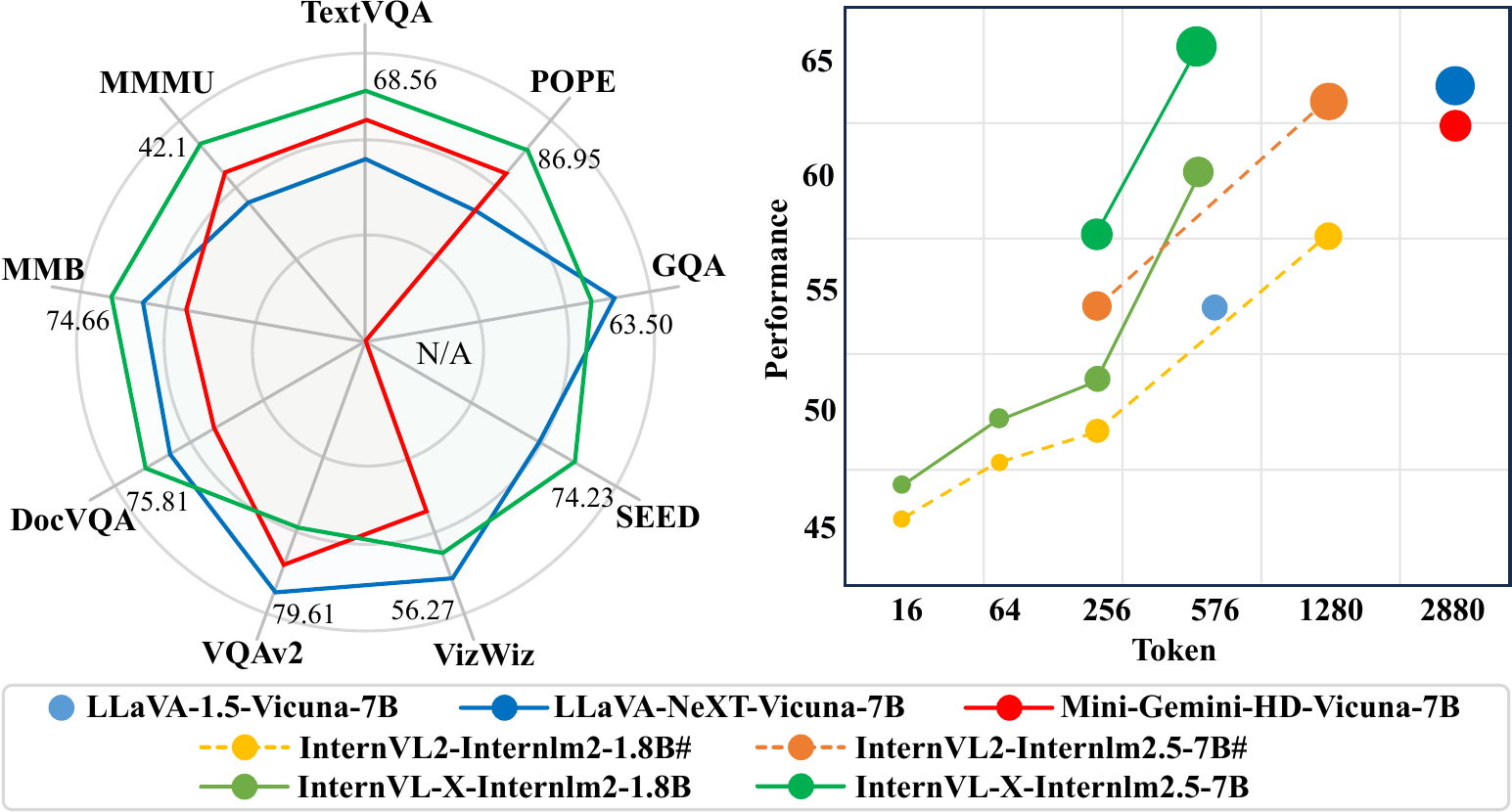}
    \caption{Comparison between InternVL-X and other models. \# indicates the model is not an official checkpoint, but a version we retrained.}  
    \label{fig1}
\end{figure}

Currently, most MLLMs adopt the LLaVA framework, where images are initially processed by a visual encoder to extract visual embeddings. These embeddings are then converted through a vision-language projector to make them interpretable by LLMs. Subsequently, LLMs process the joint representation of visual and text embeddings. Typically, LLMs have parameter capacities that exceed those of Vision Transformers (ViT) \cite{vit} by several orders of magnitude, 
thereby dominating the computational requirements in multimodal architectures. 
Additionally, the number of visual tokens can range from hundreds to thousands, contributing to significant processing delays even for simple tasks. 
Consequently, visual token compression has become an area of increasing interest for researchers.

Previous visual token compression approaches for multimodal models typically focus on optimizing individual components in isolation. 
BLIP2 and QwenVL employ learnable queries and cross-attention to extract key information from the full set of visual tokens. 
InternVL applies pixel-shuffle to merge adjacent tokens within a grid. 
FastV \cite{fastv} utilizes text-guided attention during the LLM stage to selectively retain important visual tokens. 
However, these optimization strategies fail to fully exploit the potential for model acceleration, as they overlook the opportunities of joint optimization across different components.

To address this limitation, we propose InternVL-X, which builds upon the InternVL series by incorporating three complementary visual token compression methods: Projector Visual Token Compressor (PVTC), Layer-wise Visual Token Compressor (LVTC), and Resolution Visual Token Compressor (RVTC). 
Each of these modules can independently accelerate the model while also complementing one another. 

PVTC begins by dividing visual embedding into non-overlapping grids based on its 2D space. It employs a point-to-region cross-attention that utilizes two distinct queries to extract features from the grid effectively. The local query is generated by downsampling the visual tokens within the grid, while the global query is derived from the CLS token.
This combination of queries allows the model to maintain a balanced integration of detailed local information and comprehensive global semantics.

By visualizing the attention of InternVL, we observe that most visual tokens receive low attention in the early stage of LLM, and the attention distance gradually expands as the model deepens. 
LVTC compresses tokens at the input layer and restores them at deep layers. 
To mitigate information loss during this process, we introduce multiple visual information injection paths through learnable projectors and residual connections. LVTC significantly improves the computational efficiency at different LLM stages.


Current high resolution methods, like InternVL's ratio slicing strategy, often result in computational inefficiencies when processing images with low pixel counts but extreme aspect ratios. RVTC uses the original image's pixel count or edge length to determine the optimal number of slices, effectively reducing visual token numbers. 
While maintaining a minimal decrease in performance, RVTC significantly enhances training efficiency.

Our main contributions can be summarized as follows:

1) We propose three token compression modules that effectively reduce the number of visual tokens across different MLLM stages, which significantly accelerates MLLM.

2) Through extensive experiments, we demonstrate that each module can independently improves computational efficiency. 
Moreover, combining these modules not only significantly accelerates training speed but also maintains or even enhances model performance.


3) Leveraging these modules, we present the InternVL-X model series and compare it with other MLLMs across a wide range of multimodal tasks. Our model achieves state-of-the-art results, outperforming other open-source models by a significant margin across multiple benchmarks.


\section{Related Work}

\vspace{-0.5ex}
\subsection{MLLM}
\vspace{-0.5ex}

In recent years, the field of MLLMs has rapidly evolved. Most MLLMs integrate LLMs with advanced visual encoders through a simple projector. 
BLIP series \cite{blip2,instructblip}, MiniGPT4 \cite{minigpt4}, and QwenVL \cite{qwenvl} utilize Q-Former (cross-attention) to effectively align vision-text features and extract visual information. 
Flamingo \cite{flamingo} adopts gated cross-attention to deliver encoded image information into the layers of LLMs. 
LLaVA series \cite{llava,llava_1_5,llavanext,llavaonevision} proposes a two-layer MLP structure to bridge vision and language models. 
Several subsequent works, like VILA \cite{vila}, ShareGPT4V \cite{sharegpt4v}, DeepSeekVL \cite{deepseekvl}, CogVLM \cite{cogvlm}, InternVL \cite{internvl, internvl_1_5}, etc. have followed this architecture.

\begin{figure*}[t]
    \setlength{\belowcaptionskip}{-12pt}
    \centering
    \includegraphics[width=6.85in]{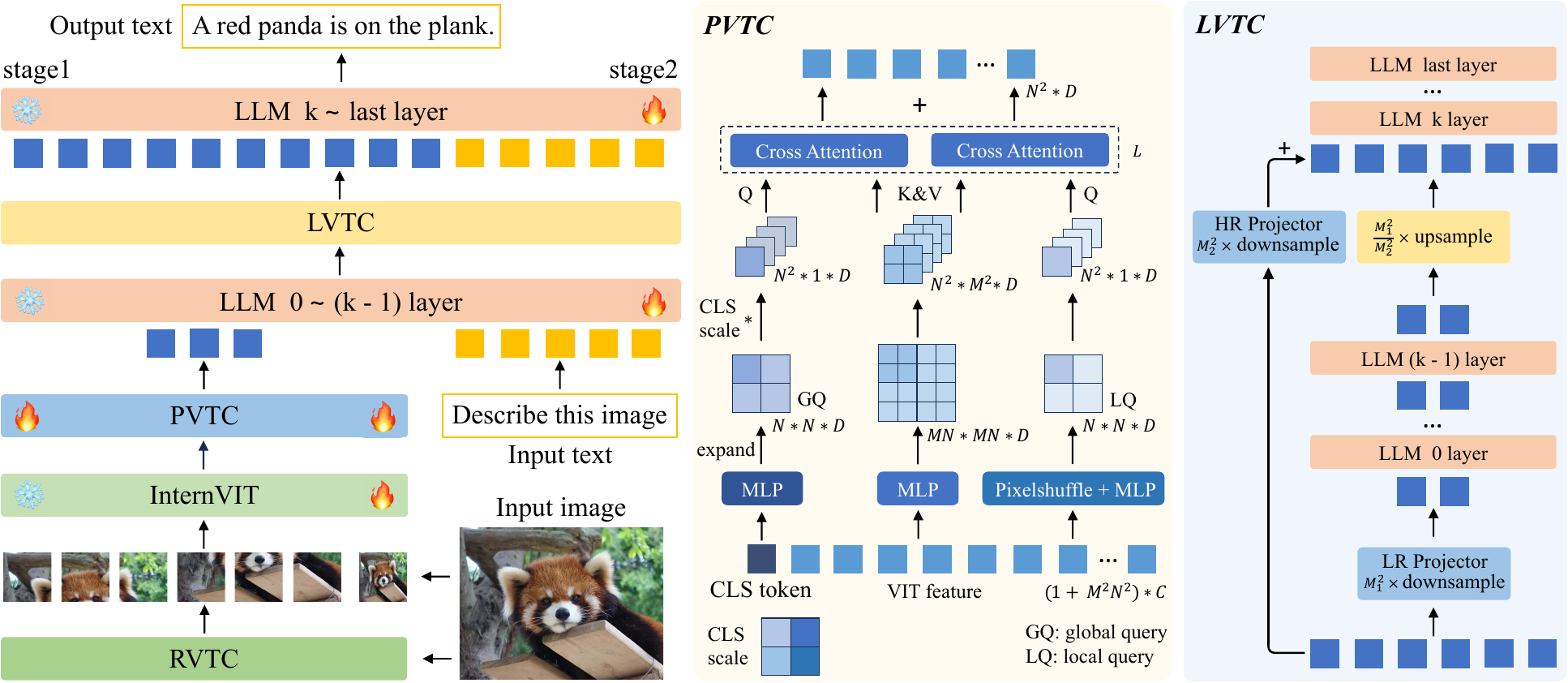}
        \caption{Architecture of the proposed InternVL-X incorporates three components: PVTC, LVTC, and RVTC. PVTC employs dual cross-attention on local and global queries to efficiently compress tokens. LVTC initially compresses visual tokens and subsequently expands them to improve their utilization across different LLM layers. RVTC optimizes image slicing to reduce the visual tokens numbers.}
        \label{fig2}
\end{figure*}

\vspace{-0.5ex}
\subsection{Visual token compression}
\vspace{-0.5ex}

Existing visual token compression methods can be broadly classified into two categories: compression during the visual projection stage, and LLM stage.

Extensive research has focused on token compression at the projection stage.
Resample \cite{qwenvl} and Q-former \cite{blip2} use learnable queries to compress token number and extract visual features through cross-attention. 
Abstractor~\cite{abstractor} and LDP \cite{mobilevlm, mobilevlmv2} adopt multi-layer convolution structures to enable local token interactions while preserving spatial relationships. 
Pixel-shuffle \cite{internvl} efficiently compresses tokens by transforming spatial dimensions into feature dimensions while preserving spatial context. 
TokenPacker \cite{tokenpacker} downsamples visual tokens to generate queries, then employs point-to-region attention to extract information from original tokens, preserving fine-grained details in the compressed tokens.
FasterVLM \cite{fastervlm} and LLaVA-PruMerge \cite{llavaprumerge} selectively retain visual tokens based on their attention scores with the CLS token.
LLaVA-Prumerge employs clustering techniques to merge discarded tokens, aiming to minimize information loss.
TRIM \cite{trim} leverages the CLIP \cite{clip} metric to evaluate the significance of each visual token. 
LLaVA-Mini~\cite{llavamini} incorporates a prefusion module to inject visual information into text tokens to mitigate performance degradation caused by compression.


Alternatively, several methods implement compression at the LLM stage.  
FastV \cite{fastv} computes the attention between vision and text tokens from the 2-nd layer of LLM and retains the top-k tokens with the highest attention scores.
FiCoco~\cite{ficoco} improves upon FastV by merging similar discarded tokens to minimize visual information loss. 
SparseVLM \cite{sparsevlm} evaluates the visual token importance through a vision-text matrix and proposes a rank-based strategy to adaptively determine the compression ratio. MustDrop~\cite{mustdrop} proposes a multi-stage compression method and comporesses visual tokens by considering both mean and maximum attention received from text tokens.
Other approaches, such as TG-Llava \cite{tgllava} and Focus-Llava \cite{focusllava}, also implement text-guided visual token compression methodologies.

\vspace{-0.5ex}    
\subsection{High resolution}
\label{sec2_3}
\vspace{-0.5ex}

High resolution is crucial for the performance of MLLMs, and many works propose their processing methods.  
LLaVA-NeXT \cite{llavanext} splits high resolution images into a grid of patches with various configurations, encodes each independently, and then merges them into a long sequence alongside the global token. 
Monkey \cite{monkey} incorporates a shared LoRA \cite{lora} structure within the visual encoder to accommodate various sub-images. 
SliME \cite{slime} utilizes adaptive slicing to scale input resolution and extracts contextual information from the global view using a mixture of adapters.
Mini-Gemini-HD \cite{minigemini} employs dual encoders: the low resolution generates visual queries, while the high resolution provides keys and values, utilizing attention for enhanced fine-grained feature extraction. 
InternVL \cite{internvl_1_5} and Internlm-Xcomposer2-4KHD \cite{Internlm_xcomposer2_4khd} further expands the dynamic multi-resolution range, supporting up to 4K HD images, significantly improves performance.


\section{Method}

\textbf{Overview.} In this section, we first introduce the main bottlenecks of current MLLMs, and then present our method.

More visual tokens provide richer visual information and significantly improve the model performance, but due to the $n^2$ complexity of the transformer \cite{transformer} model, a large number of visual tokens cause explosive growth in computational complexity. During the forward process of the MLLM, visual tokens typically exceed text tokens by over tenfold, which seriously affects the training efficiency. Balancing the performance and efficiency of MLLMs is an urgent issue that needs to be addressed.



In this work, we introduce InternVL-X, an enhanced version of the InternVL model. The architecture of InternVL-X, depicted in Fig.~\ref{fig2}, comprises a visual encoder, a projector, and an LLM. Given an image $V \in R^{H * W * 3}$, the visual encoder transforms it into CLS token $F_{CLS} \in R^{1 * C}$ and a set of visual embeddings $F_{V} \in R^{S^2 * C}$. 
The projector then scales down $F_{V}$ by a factor of $M^2$ to produce the compressed visual token $F_{C} \in R^{N^2 * D}$, where $N = S / M$. Finally, we concatenate $F_{C}$ with text token $F_{T}$ and feed them into the LLM to generate the output via an autoregressive paradigm. Please note that in this paper, unless otherwise specified, all $\times$ symbols represent multiples and all $*$ symbols represent multiplications.

We mainly compress visual tokens from the model and data perspectives. At the model level, we propose a new projector PVTC and an LLM compression module LVTC. PVTC serves as a bridge between the visual encoder and LLM, effectively converting visual features through global and local cross-attention mechanism. 
LVTC utilizes fewer visual tokens in the early LLM layers and only increases the tokens in the last few layers to enhance efficiency.
At the data level, we design a high resolution slicing method RVTC, which reduces the number of slices by filtering the image area or edge length. 
Next, we will introduce each module separately.

\subsection{PVTC}



Existing projector token compression methods, like InternVL and MobileVLM, utilize pixel-shuffle or convolutional architectures to reduce visual tokens, often sacrificing critical local details in the process. TokenPacker partially addresses this limitation by employing cross-attention with downsampled local queries, but it neglects the global semantic information.

To address this, we propose PVTC, which leverages cross-attention mechanisms between visual tokens and dual queries—derived from local windows and the global CLS token, enabling efficient visual compression while preserving both fine-grained local details and comprehensive global semantics.

Specifically, our approach begins by recovering the spatial organization of visual embeddings from the final visual encoder layer. 
We first reshape the flattened visual embeddings $F_{V} \in R^{S^2 * C}$ back into their original 2D spatial structure $F_{V} \in R^{S * S * C}$. 
These embeddings are then partitioned into non-overlapping $N * N$ windows, each with $M*M$ tokens, where $M=2^{i},i \in{[0, \log_{2}{S}]}$. 
Within each window, we employ two parallel paths to generate the final compressed visual embeddings.


\begin{figure}[h]
  \setlength{\abovecaptionskip}{12pt}
  \setlength{\belowcaptionskip}{-8pt}
  \centering
  \hspace{-2.8ex}
  \begin{subfigure}{0.23\linewidth}
      \includegraphics[height=1.05in]{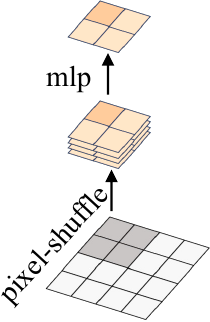}
      \caption{Pixel-shuffle}
  \end{subfigure}
  \hspace{-1.3ex}
  \begin{subfigure}{0.23\linewidth}
      \includegraphics[height=1.05in]{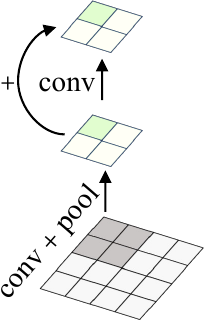}
      \caption{LDPv2}
  \end{subfigure}
  \hspace{-1.8ex}
  \begin{subfigure}{0.25\linewidth}
      \includegraphics[height=1.05in]{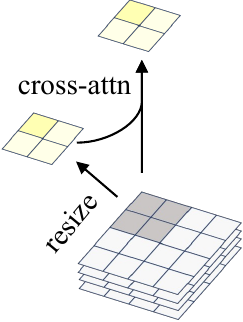}
      \caption{TokenPacker}
  \end{subfigure}
  \hspace{-1.1ex}
  \begin{subfigure}{0.25\linewidth}
      \includegraphics[height=1.05in]{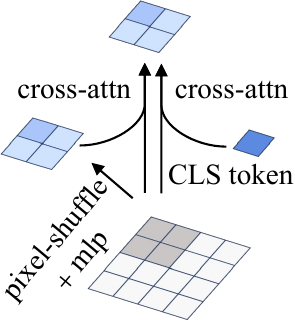}
      \caption{PVTC}
  \end{subfigure}
  \caption{Comparison of different projectors.}
  \label{fig3}
\end{figure}

\textbf{Local Attention.} 
For all visual tokens within a window, we compress them into 1 token through a pixel-shuffle operation, which transposes the $M^2$ tokens in each window from the spatial to feature dimension.
These compressed $N * N$ tokens are then transformed through an MLP layer to match the feature dimension of the LLM, generating the local embedding $F_{LQ} \in R^{N * N * D}$. 
Subsequently, we convert the original visual embedding $F_{V}$ to the same feature dimension to obtain $F_{KV} \in R^{S * S * D}$. 
$F_{LQ}$ serves as the local query in cross-attention while $F_{KV}$ acts as key and value. 
To reduce the amount of computation, we adopt the point-to-region cross-attention in \cite{tokenpacker}, where each query only calculates the attention within its corresponding window. We accomplish this through a dimensionality transformation, and get $F_{LQ} \in R^{N^2 * 1 * D}$, $F_{KV} \in R^{N^2 * M^2 * D}$.
By applying cross-attention to each window, we produce compressed token $F_{L} \in R^{N * N * D}$, which enhances the retention and integration of local embeddings.

\textbf{Global Attention.} We utilize the CLS token as the global query, which aggregates the global information of the image and could effectively extract the important information. We first transform the feature dimension of $F_{CLS}$ from $C$ to $D$ through an MLP layer, and then expand it to 2D space to get $F_{GQ} \in R^{N * N * D}$. To avoid extracting much redundant information from similar tokens, we introduce diverse transformations to $F_{GQ}$ at each position. Specifically, we set a learnable scale vector CLS scale $\in R^{N * N * D}$, which is multiplied by $F_{GQ}$ to obtain a diverse global query. Similarly, $F_{GQ}$ interacts with $F_{KV}$ in each window separately. Through this process, we get global compressed embeddings $F_{G} \in R^{N * N * D}$, effectively capturing the overall semantic information.

After extracting features through $L$ layers of attention, we combine $F_{L}$ and $F_{G}$ through element-wise addition and then reshape the result back into a 1D format to obtain the final compressed embeddings $F_{C} \in R^{N^2 * D}$. This combined representation retains both localized spatial details and comprehensive contextual information, achieving efficient compression while mitigating information loss.

\subsection{LVTC}

\begin{figure}[h]
  \setlength{\abovecaptionskip}{4pt}
  \centering
  \begin{subfigure}{0.36\linewidth}
    \centering
      \includegraphics[height=0.72in]{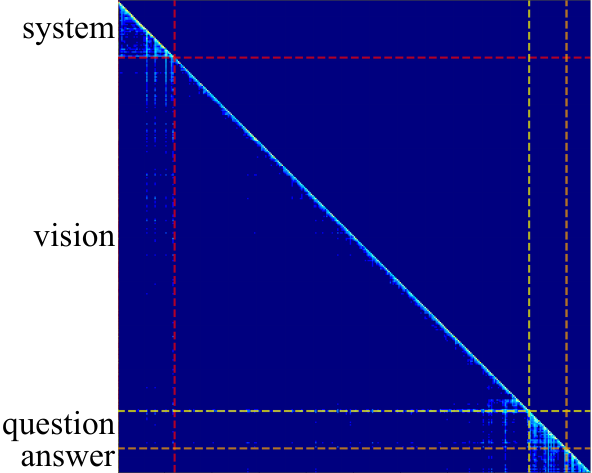}
      \caption{layer0}
  \end{subfigure}
  \hspace{-0.8ex}
  \begin{subfigure}{0.3\linewidth}
    \centering
      \includegraphics[height=0.72in]{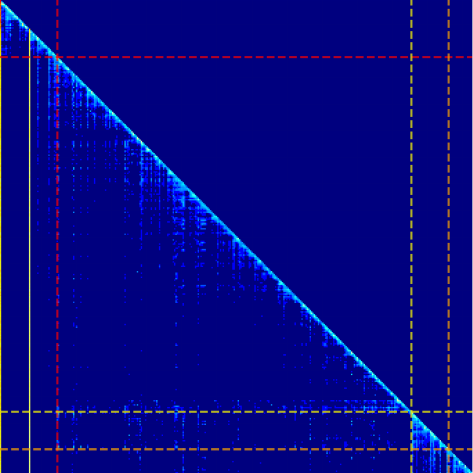}
      \caption{layer5}
  \end{subfigure}
  \hspace{-0.95ex}
  \begin{subfigure}{0.32\linewidth}
    \centering
      \includegraphics[height=0.72in]{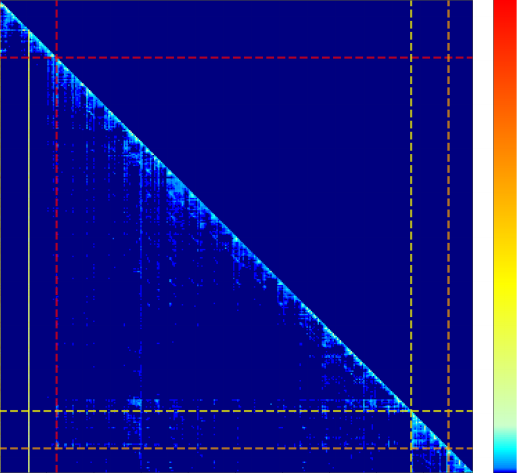}
      \caption{layer11}
  \end{subfigure}
  \begin{subfigure}{0.36\linewidth}
    \centering
      \includegraphics[height=0.72in]{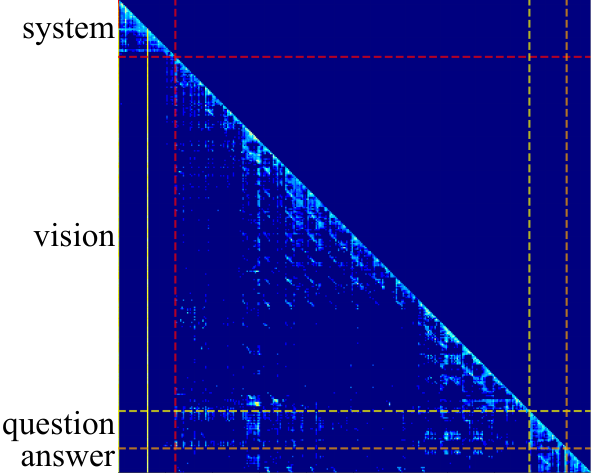}
      \caption{layer15}
  \end{subfigure}
  \hspace{-0.8ex}
  \begin{subfigure}{0.3\linewidth}
    \centering
      \includegraphics[height=0.72in]{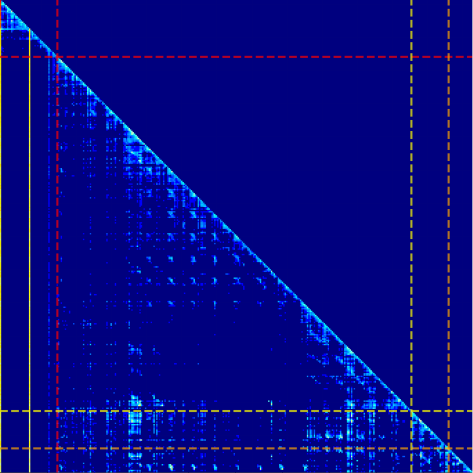}
      \caption{layer19}
  \end{subfigure}
  \hspace{-0.95ex}
  \begin{subfigure}{0.32\linewidth}
    \centering
      \includegraphics[height=0.72in]{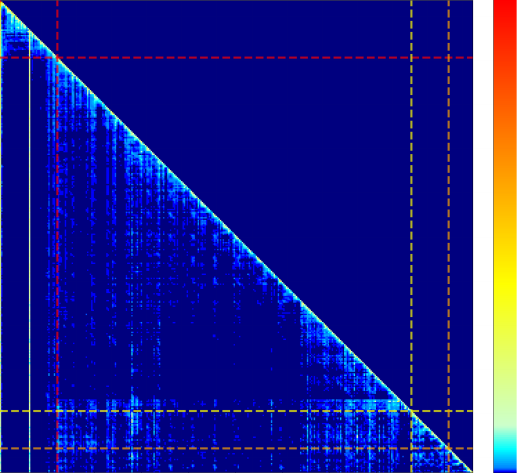}
      \caption{layer23}
  \end{subfigure}
  \caption{Viualization of average attention weights of each token in the LLM process. The horizontal axis is the key position and the vertical axis is the query position.}
  \label{fig4}
\end{figure}

Previous studies, such as FastV and LLaVA-Mini, observe that most visual tokens receive high attention in the early layers of LLM, with this attention decreasing in deeper layers. However, the InternVL model demonstrates the opposite behavior. Our analysis, shown in Fig. \ref{fig4}, reveals that in the initial layers, both visual and text tokens mainly focus on their adjacent areas. As the model progresses to the deeper layers, the attention distance gradually increases. In the middle layers, only about $1/3$ the visual tokens are activated by text tokens. In the final layers, most tokens gain high attention, peaking in the last layer. This suggests that in the early stages, attention is primarily distance-based. Consequently, numerous visual tokens in the early layers of LLM may not enhance performance and could undermine computational efficiency.

\vspace{-0.5ex}

\vspace{-2ex}

\begin{figure}[h]
  \vspace{1.2ex}
  \setlength{\abovecaptionskip}{8pt}
  \setlength{\belowcaptionskip}{-14pt}
  \centering
  \includegraphics[width=3in]{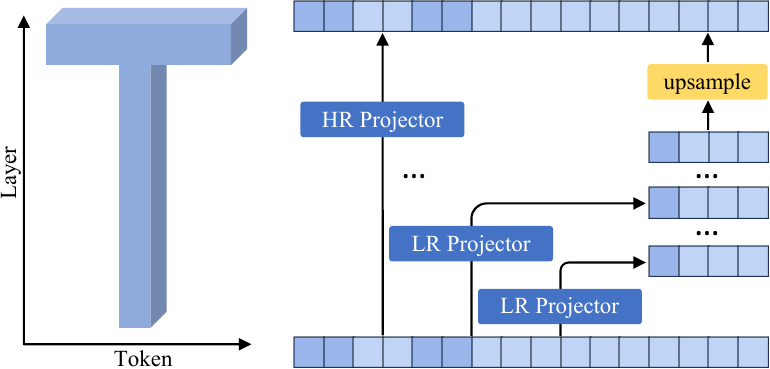}
    \caption{LVTC uses a high resolution projector and multi-projector structure to enhance visual information in LLM.} 
    \label{fig5}
\end{figure}

Building on the above observations, we design a layer-wise visual token compression module LVTC. 
Unlike most previous methods that progressively compress tokens layer by layer, our approach compresses tokens right at the input layer and increase tokens at a specific layer $k$. 
During the forward pass of LLM, the token presents a ``T" shape, as illustrated in Fig. \ref{fig5}. 
In detail, we first perform a low resolution (LR) compression, leveraging an LR projector to compress $F_{V}$ by a factor of $M_{1}^2$, thus obtaining $N_{1}^2$ tokens, where $M_{1} = S / N_{1}$. 
Then we calculate attention on these $N_{1}^2$ tokens in the first $k$ layers of LLM.
At layer $k$, we expand the token count to $N_{2}^2$ through an Upsample2D process, with an upsampling scale factor of $M_{1}^2 / M_{2}^2$, where $M_{2} = S / N_{2}$, $M_{1} > M_{2}$ and $N_{2} \gg N_{1}$.
This methodology allows most computations to focus on the reduced set of $N_{1}^2$ tokens across most layers, only increases to $N_{2}^2$ tokens in the final layers.
Compared to using $N_{2}^2$ tokens throughout all layers, our approach maintains identical performance while significantly decreasing computational demands. 

\begin{algorithm}[h]
  \footnotesize
  \caption{LVTC}
  \label{alg1}
  \renewcommand{\algorithmicendfor}{\relax} 
  \renewcommand{\algorithmicendif}{\relax}
  \begin{algorithmic}
  \STATE \# V: Visual token, T: Text token, s: insert start layer, i: insert interval, k: token expansion layer, 
  vis\_pos: position of visual token,  LRProjs, HRProj, VLR, VHR\\
  \vspace{1ex}
  \FOR {proj in LRProjs}
  \STATE VLR.append(proj(V))
  \vspace{-2.5ex}
  \ENDFOR
  \STATE VHR = HRProj(V)
  \STATE X = concat(VLR[0], T); del VLR[0]

  \FOR {idx, layer in enumerate(LLM.layers)}
  \IF {idx == k}
  \STATE X = concat((upsample2d(X[vis\_pos]) + VHR, X[!vis\_pos])
  \ELSIF {s $<$= idx $<$ s + i * len(VLR) and (idx - s) \% i == 0}
  \STATE X[vis\_pos] += VLR[(idx - s) // i]
  \ENDIF
  \vspace{-2.5ex}
  \STATE X = layer(X)
  \vspace{-2.5ex}
  \ENDFOR
  \end{algorithmic}
\end{algorithm}
\vspace{-1ex}


Although LVTC expands visual tokens in deep layers, the upsampling operation does not bring more visual information.
Inspired by ResNet \cite{resnet} and FPN \cite{fpn}, we introduce an additional high resolution (HR, different from the concept in Sec. \ref{sec2_3}) compression that converts $F_{V}$ into $N_{2}^2$ tokens through a HR projector. These tokens are then integrated with the upsampled $N_{2}^2$ tokens at layer $k$ through residual connections. This approach effectively injects visual information into the LLM without loss.
To further diversify the visual input and mitigate the risk of visual knowledge forgetting, we incorporate a multi-projector strategy into the middle layers of LLM through residual connections. 
Specifically, we set $T$ LR projectors to obtain $T$ groups of compressed tokens. Beginning at the $s-th$ layer of LLM, we integrate a set of compressed tokens into the visual tokens at intervals of $i$ layers.
The LVTC structure and computational process are shown in Fig. \ref{fig5} and Alg. \ref{alg1}, which enhances the model’s capacity to retain and utilize visual information effectively.

\subsection{RVTC}

Several models, like LLaVA-NeXT and InternVL, use high resolution strategy to enhance performance. However, this approach leads to a significant increase in visual tokens and computational demands. 
For instance, setting the number of slices to 4 will result in a 2.5$\times$ increase in training time.
InternVL predefines several aspect ratios and determines the number of slices by matching the optimal ratio. 
Although effective, this method can cause computational inefficiencies, particularly when processing small images with large aspect ratios. To illustrate, if the maximum number of patches is set to 6, an $80*20$ image will be assigned a 4:1 ratio and divided into 4 patches, while a $345*372$ image fits best with a 1:1 ratio and remains unsliced. This ratio matching mechanism results in an uneven distribution of computational resources, significantly affecting training efficiency.

\vspace{-1.7ex}
\begin{figure}[h]
  \setlength{\abovecaptionskip}{8pt}
  \setlength{\belowcaptionskip}{-6pt}
  \centering
  \includegraphics[width=3.2in]{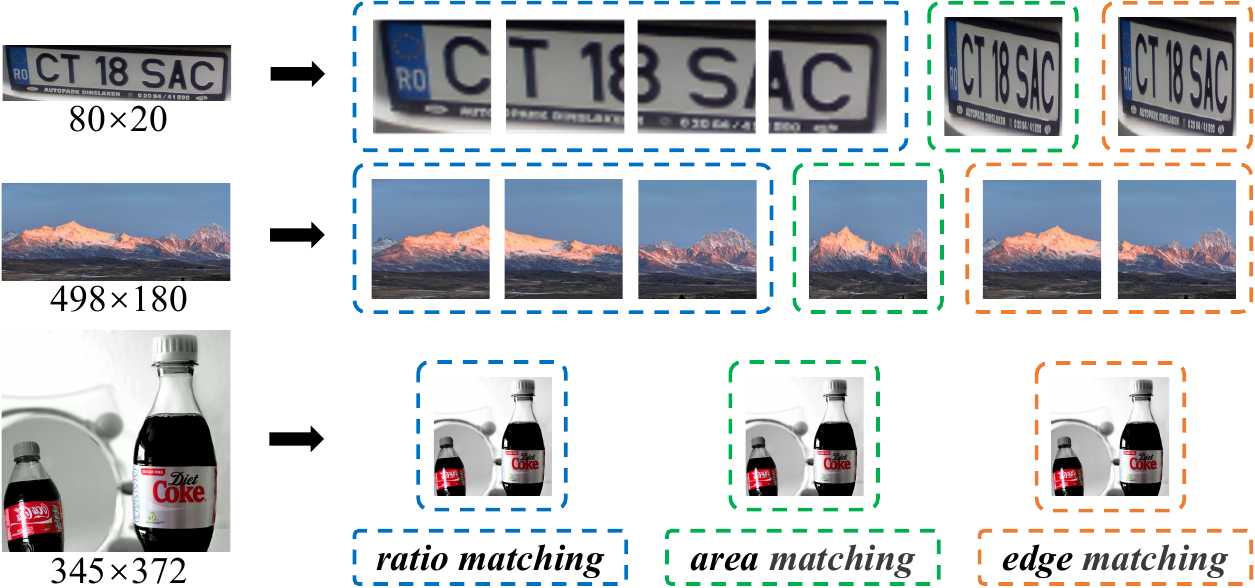}
    \caption{Differnt slicing methods. Compared to ratio matching, area matching and edge matching require fewer slices.} 
    \label{fig6}
\end{figure}

\begin{table*}[t]
    \setlength{\abovecaptionskip}{8pt}
    \setlength{\belowcaptionskip}{-10pt}

    \centering

    {
    \vspace{-1ex}
    \resizebox{\linewidth}{!}
    {
    \begin{tabular}{@{}l@{\hskip 0.5em}|@{\hskip 0.5em}l@{\hskip 0.5em}|@{\hskip 0.5em}l@{\hskip 0.5em}|@{\hskip 0.5em}l@{\hskip 0.5em}|@{\hskip 0.5em}l@{\hskip 0.5em}l@{\hskip 0.5em}l@{\hskip 0.5em}l@{\hskip 0.5em}|@{\hskip 0.5em}l@{\hskip 0.5em}l@{\hskip 0.5em}l@{\hskip 0.5em}|@{\hskip 0.5em}l@{\hskip 0.5em}l@{\hskip 0.5em}l@{\hskip 0.5em}l@{\hskip 0.5em}l@{\hskip 0.5em}|@{\hskip 0.5em}l@{}}

    \toprule 
    \multirow{3}{*}[0pt]{\makecell[c]{Model}} & 
    \multirow{3}{*}[0pt]{\makecell[c]{LLM}} & 
    \multirow{3}{*}[0pt]{\makecell[c]{PT/IT}} &
    \multirow{3}{*}[0pt]{\makecell[c]{Token}} & 
    \multicolumn{4}{c|@{\hskip 0.5em}}{Text-oriented VQA} & 
    \multicolumn{3}{c|@{\hskip 0.5em}}{General VQA} & 
    \multicolumn{5}{c|@{\hskip 0.5em}}{Comprehensive benchmark} &
    \multirow{3}{*}[0pt]{\makecell[c]{Avg}} \\ 
                &                                    &                         &         &Text     &Doc     &Chart     &Info      &GQA     &VQA     &Viz     &MMB     &MM       &MM      &PO     &SEED\\
                &                                    &                         &         &VQA      &VQA     &QA        &VQA       &        &v2      &Wiz     &        &Vet      &MU      &PE     &-IMG\\
    \midrule
    
    MobileVLM V2\cite{mobilevlmv2}      &Mobilellama-2.7B      &1.2M/3.6M      &144      &52.1     &-       &-         &-         &59.3    &-       &-       &-       &-        &-       &84.3   &-       &-\\ 
    BLIP-2\cite{blip2}                  &Vicuna-13B            &129M/-         &32       &42.5     &-       &-         &-         &41.0    &65.0    &19.6    &-       &-        &-       &85.3   &49.7     &-\\
    Insturct-BLIP\cite{instructblip}    &Vicuna-7B             &129M/1.2M      &64       &50.1     &-       &-         &-         &49.5    &-       &34.5    &-       &26.3     &-       &-      &-     &-\\
    QwenVL\cite{qwenvl}                 &Qwen-7B               &1.4B/50M       &256      &63.8     &65.1*   &65.7*     &-         &59.3*   &78.8*   &35.2    &-       &-        &-       &-      &62.3     &-\\
    VILA\cite{vila}                     &Llama2-7B             &50M/1M         &576      &64.4     &-       &58.6*     &-         &62.3*   &79.9*   &57.8    &68.9    &34.9     &-       &85.5   &-        &-\\
    MobileVLM V2\cite{mobilevlmv2}      &Vicuna-7B             &1.2M/3.6M      &144      &62.3     &-       &-         &-         &62.6    &-       &-       &-       &-        &-       &85.3   &-       &-\\
    MiniGemini\cite{minigemini}         &Vicuna-7B             &1.2M/1.5M      &576      &65.9     &-       &-         &-         &-       &-       &-       &68.5    &46.0     &38.1    &-      &-        &-\\
    LLaVA-1.5\cite{llava_1_5}           &Vicuna-7B             &558K/665K      &576      &58.2     &28.1    &-         &25.8      &63.3*   &78.5*   &50.0    &64.3    &31.1     &35.3    &85.9   &66.1     &-\\
    TokenPacker\cite{tokenpacker}       &Vicuna-7B             &558K/665K      &144      &-        &-       &-         &-         &61.9*   &77.9*   &52.0    &65.1    &33.0     &-       &87.0   &-        &-\\
    InternVL2\cite{internvl_1_5}\#      &Internlm2-1.8B        &558K/665K      &256      &44.99    &23.21   &14.23     &16.56     &61.03*  &74.20*  &47.55   &59.75   &24.6     &29.7    &86.33  &64.48    &45.47\\
    InternVL2\cite{internvl_1_5}\#      &Internlm2.5-7B        &558K/665K      &256      &49.73    &26.94   &18.08     &21.82     &63.01*  &77.79*  &50.56   &70.85   &34.1     &39.2    &86.78  &71.13    &50.83\\

    \midrule
    InternVL-X                          &Internlm2-1.8B        &558K/665K      &256      &49.91    &27.31   &19.03     &17.71     &61.55   &74.86*  &51.83   &63.34   &28.6     &32.2    &86.21  &65.39    &48.32\\
    InternVL-X                          &Internlm2.5-7B        &558K/665K      &256      &55.24    &31.72   &21.83     &23.90     &63.62   &78.69*  &54.51   &72.64   &41.7     &41.9    &87.13  &72.58    &53.85\\

    \midrule
   
    Monkey\cite{monkey}                 &Qwen-7B            &-/1.44M      &$\sim$1024   &67.7     &66.5    &36.1      &-        &60.7    &80.3*   &61.2    &-       &-        &-       &-      &-       &-\\
    TokenPacker-HD\cite{tokenpacker}    &Vicuna-7B          &1.2M/1.5M    &$\sim$954    &68.0     &60.2    &-         &-         &-       &81.2*   &54.7    &67.4    &-        &35.4    &-      &-       &-\\
    Mini-Gemini-HD\cite{minigemini}     &Vicuna-7B          &1.2M/1.5M    &2880         &68.4     &65.0    &-         &-         &-       &80.3*   &54.6    &65.8    &41.3     &36.8    &86.8   &-       &-\\
    LLaVA-UHD\cite{llavauhd}            &Vicuna-13B         &595K/665K    &$\sim$256    &67.7     &-       &-         &-         &-       &81.7*   &56.1    &-       &-        &-       &89.1   &-       &-\\
    LLaVA-NeXT\cite{llavanext}          &Vicuna-7B          &558K/765K    &$\sim$2880   &64.9     &74.4    &54.8      &37.1      &64.2    &81.8*   &57.6    &68.1    &43.9     &35.8    &86.5   &68.2    &61.38\\
    InternVL2-HD\cite{internvl_1_5}\#   &Internlm2-1.8B     &558K/770K    &$\sim$1282   &59.59    &66.60*  &62.80*    &25.69     &60.24*  &75.60*  &51.24   &62.50   &27.6     &29.1    &87.36  &66.93   &56.27\\
    InternVL2-HD\cite{internvl_1_5}\#   &Internlm2.5-7B     &558K/770K    &$\sim$1282   &65.58    &72.64*  &69.77*    &30.87     &63.24*  &78.85*  &56.33   &72.08   &35.7     &39.9    &87.25  &73.42   &62.13\\                         

    \midrule
    InternVL-X-HD                       &Internlm2-1.8B     &558K/770K    &$\sim$520   &63.32    &69.22*  &64.48*    &28.75     &61.35*  &76.52*  &57.24   &63.67   &30.8     &33.2    &87.25  &67.72   &58.64\\
    InternVL-X-HD                       &Internlm2.5-7B     &558K/770K    &$\sim$520   &68.56    &75.81*  &73.24*    &38.05     &63.50*  &79.61*  &56.27   &74.66   &40.7     &42.1    &86.95  &74.23   &64.47\\
    \bottomrule
    \end{tabular}
    }
    }
    \caption{Comparison of Different Methods on 12 benchmarks. * indicates that the training set of this benchmark is visible during the model training. \# means that the model is not an official checkpoint, but our reproduced version.}
    \label{table1}
\end{table*}

To tackle this issue, we introduce an efficient image slicing method RVTC. The amount of information an image is largely determined by its pixel count. Building on this principle, we propose two innovative matching strategies: area matching and edge length matching. Consider an image with a resolution of $(W, H)$, and an input resolution for the MLLM set at $(448, 448)$, along with a predefined maximum number of patches $R=6$. In the area matching method, we take $448*448$ as a basic unit to determine the maximum number of slices. Specifically, we first compute $R=min(R, \lceil W*H/(448*448) \rceil)$, then execute the optimal match. Similarly, in the edge length matching method, we utilize the side length as the measurement unit to calculate $R$. In this case, $R=min(R, R_{W}*R_{H})$, where $R_{W}=\lceil W/448 \rceil$, $R_{H}=\lceil H/448 \rceil$. Fig. \ref{fig6} illustrates a comparison of different slicing methods, highlighting the completeness and efficiency of RVTC in image slicing. 
This strategy maximizes the utilization of available pixels, thereby enhancing the efficiency of model training.


\section{Experiments}

This section presents our experimental setup and results. We first describe our model training specifications, followed by evaluation benchmarks and datasets. We then compare our model's performance with existing methods and conclude with ablation studies on the proposed methods.

\subsection{Implement details and datasets}
\textbf{Experiments configuration.} We initialize InternVL-X models with the same pre-trained components as the InternVL2 series. For the 2B model, we employ InternViT-300M-448px \cite{internvl} as the visual encoder and InternLM2-Chat-1.8B \cite{internlm2} as the LLM, while the 8B variant uses the same visual encoder but upgrades to InternLM2.5-Chat-7B \cite{internlm2_5} for LLM. Our training strategy follows a two-stage approach similar to LLaVA-1.5. In the pretraining stage, we finetune only the projector module and freeze other parameters, followed by a instruction tuning stage where all model parameters are finetuned. All models are trained for 1 epoch using the AdamW optimizer with a cosine learning rate scheduler. The initial learning rate is set to 4e-5. The training is conducted on 8 NVIDIA A800 GPUs. In normal resolution experiments, we use PVTC and LVTC, and in high resolution, we adopt PVTC, LVTC, and RVTC.

\textbf{Training datasets.} We follow the training datasets as LLaVA-1.5 and LLaVA-NeXT. In the first stage, we utilize LCS-558K \cite{llava} dataset to finetune the projector. In the second stage, for normal resolution model, we use LLaVA-665K \cite{llava} dataset for instruction tuning. For high resolution model, as the dataset used in LLaVA-NeXT is not publicly available, we assemble an instruction tuning dataset with 770K samples following the guidance of LLaVA-NeXT.

\textbf{Evaluation benchmarks.} For fair evaluation, we assess our model on 12 diverse benchmarks, which span text-oriented VQA (TextVQA \cite{textvqa}, DocVQA \cite{docvqa}, ChartQA \cite{chartqa}, InfoVQA \cite{infovqa}), general VQA (GQA \cite{gqa}, VQAv2 \cite{vqav2}, VizWiz \cite{vizwiz}), and comprehensive evaluation (MMB \cite{mmb}, MMVet \cite{mmvet}, MMMU \cite{mmmu}, POPE \cite{pope}, SEED \cite{seed}).

\vspace{-0.5ex}
\subsection{Quantitive results}
\vspace{-0.5ex}

We conduct comprehensive evaluations of our InternVL-X series models across 12 benchmarks. 
To ensure fair comparison, we implement InternVL2 by training the official InternVL2 architecture from scratch, using the same training data and initialization as our InternVL-X, rather than using the official InternVL2 checkpoints.
As shown in Tab. \ref{table1}, compared with InternVL2, the 2B and 8B versions of InternVL-X improve the average performance by 2.85\% and 3.02\% across all 12 benchmarks, respectively. With high resolution, InternVL-X improves these two metrics by 2.37\% and 2.34\%. 
Futhermore, InternVL-X-8B achieves the state-of-the-art results on 7 benchmarks. Using less than 20\% of the tokens, we exceed the average performance of LLaVA-NeXT by 2.34\%.

\textbf{Text-oriented VQA.} 
We evaluate InternVL-X against other models across 4 challenging benchmarks. Our 2B and 8B models demonstrate an average improvement of 4.22\% and 4.23\%, respectively, over the baseline model InternVL2. Notably, some tasks like DocVQA are zero-shot evaluations. Our 8B version achieves best result on the DocVQA task.
In the TextVQA and InfoVQA tasks, despite our models showing considerable improvements, there remains a gap between our model and those of SOTA models like LLaVA-1.5. 
In high resolution evaluation, more training data and finer-grained visual information lead to substantial improvements. Impressively, the InternVL-X-8B achieves the best results in all of the 4 tasks.

\textbf{General VQA.} 
We conduct an evaluation on 3 classic tasks: VQAv2, GQA, and VizWiz. As presented in Tab. \ref{table1}, InternVL-X surpasses the baseline model across all 3 tasks, enhancing the average metric by 2.15\% and 1.82\%. 
Our model excels particularly on the GQA task, achieving the best performance. 
While our approach slightly lags behind the leading models on the VQAv2 and VizWiz tasks, it still delivers competitive results.

\textbf{Comprehensive benchmark.} 
Our approach demonstrates exceptional performance across the MMB, MMMU, and SEED-IMG benchmarks. In high resolution evaluations, InternVL-X surpasses LLaVA-NeXT, improving the MMB and MMMU metrics by 6.56\% and 6.3\%, respectively. 
Unfortunately, our method is slightly inferior to the baseline on the hallucination benchmark POPE. We speculate that richer visual information may increase the probability of hallucination in the model. 
Impressively, our 2B model could achieve comparable results to other 7B models in certain metrics, showcasing the outstanding performance of InternVL-X on visual benchmarks.

\subsection{Ablation study}
\vspace{-0.5ex}
We conduct sufficient ablation experiments on our proposed method using the InternVL2 model as the base model. If there is no special explanation, the following ablation experiments are all conducted on the 2B model with normal resolution.  We report the performance on 6 benchmarks, including TextVQA, DocVQA, GQA, VizWiz, MMVet and POPE. Next, we will introduce each part in detail.

\begin{table}[h]
    \vspace{1.2ex}
    \centering
    \setlength{\abovecaptionskip}{4pt}
    \setlength{\belowcaptionskip}{-8pt}
    \resizebox{0.9\linewidth}{!}
    {
        \begin{tabular}{@{}l@{\hskip 1em}l@{\hskip 1em}ll@{}}
        \toprule
        Method         & Token               & Accuracy    & Train/Inference Time \\
        \midrule
        Base           & 256                 &47.78        &4.1h/39.9ms       \\ 
        Base*          & 1024                &49.93        &11.1h/40.6ms     \\
        Base-HD   & $\sim$1282               &58.77        &13.5h/46.7ms      \\
        \midrule
        + PVTC         & 256                 &50.03        &4.3h/42.7ms   \\
        + LVTC*        & 256                 &50.26        &7.0h/40.3ms   \\ 
        + PVTC + LVTC* & 256                 &50.90        &7.3h/41.6ms   \\
        + RVTC         & $\sim$520          &58.21        &8.6h/43.3ms   \\
        + PVTC + RVTC  & $\sim$520          &60.85        &9.1h/44.5ms   \\
        Full Model*    & $\sim$520          &61.56        &18.7h/45.2ms   \\
        \bottomrule
        \end{tabular}
    }
    \caption{Ablation study of different components. * indicates that flash attention is not used. The training time in the table is the sum of PT and IT time, and the inference time is the average generation latency of per token.}
    \label{table2}
\end{table}

We explore the complementary interactions among our three proposed modules and their individual impacts on models, as illustrated in Tab. \ref{table2}. PVTC alone improves the model's performance by 2.25\% across 6 benchmarks. 
With more visual information, LVTC could also significantly improve the model effect, and it could bring superposition of effects with PVTC. 
However, However, as LVTC has not yet been implemented on flash attention \cite{flashattention}, it has somewhat increased the training time. Nevertheless, compared to the baseline experiment without flash attention and token compression, LVTC demonstrates notable enhancements in both performance and efficiency.
Although RVTC slightly diminishes model performance, it achieves considerable reductions in training time.
Moreover, when all three modules are combined into a single model, both performance and efficiency exceed those of models utilizing individual modules, underscoring the complementary nature of our proposed modules.  
Additionally, since kv-cache is used to accelerate the inference process, we find that the impact of token length on inference time is almost negligible.

\begin{figure}[h]
    \setlength{\abovecaptionskip}{4pt}
    \setlength{\belowcaptionskip}{-8pt}
    \centering
    \includegraphics[width=3.25in]{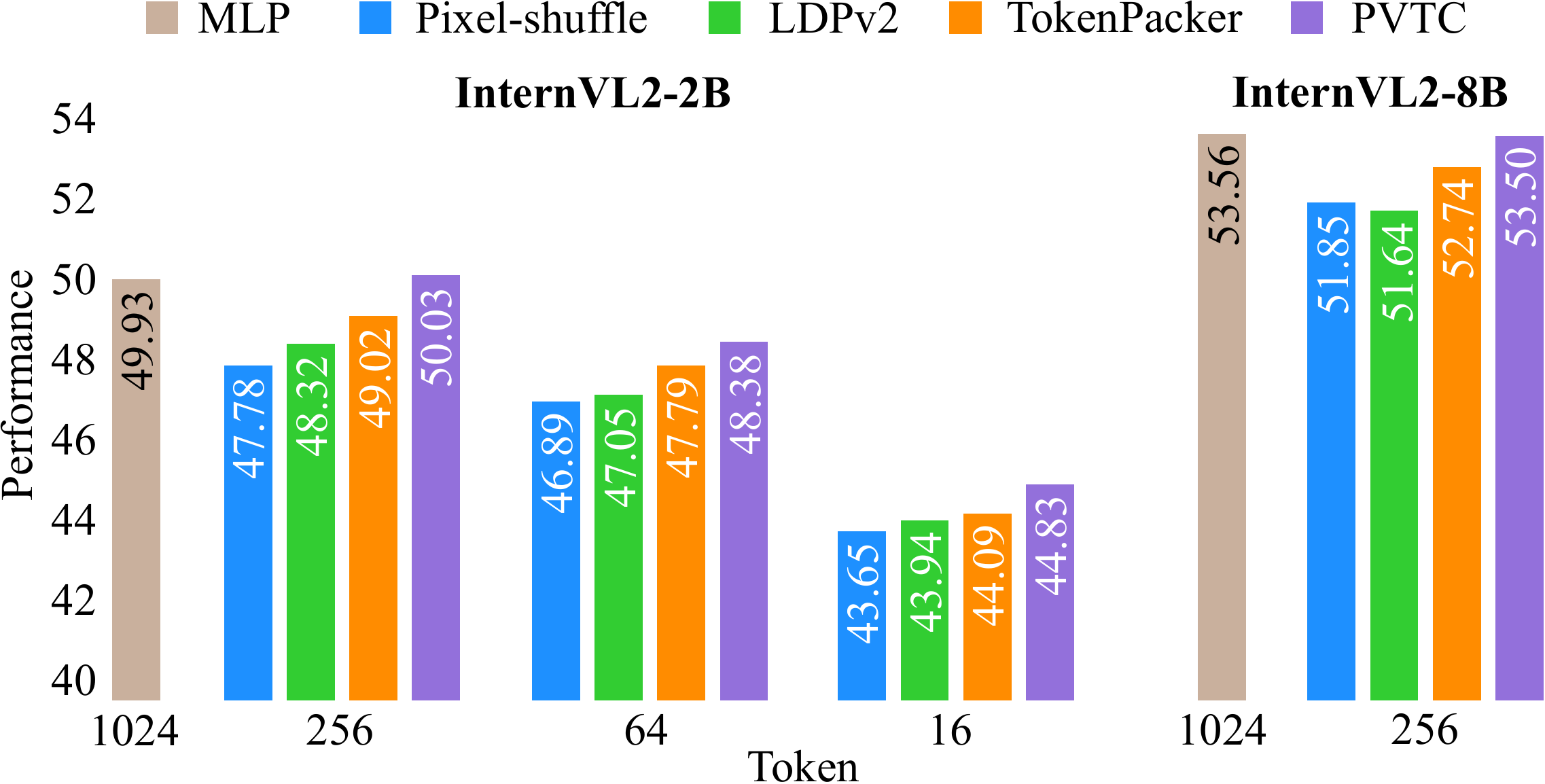}
    \caption{Comparison of various projectors under different compression ratios.
    } 
    \label{fig7}
\end{figure}

\textbf{PVTC.}
To validate the effectiveness of our proposed PVTC module, we compare PVTC with other projectors, including Pixel-shuffle \cite{internvl}, LDPv2 \cite{mobilevlmv2}, and TokenPacker \cite{tokenpacker}. For a fair comparison, we utilize different projectors on InternVL2 to replace the original projector, and all methods are trained with the same training data and experimental settings. We first perform linear projector(MLP) \cite{llava_1_5} to assess model performance without token compression (1024 token). Subsequently, we conducte experiments with various projectors under 4$\times$, 16$\times$, and 64$\times$ compression ratios (corresponding to 256, 64, and 16 tokens, respectively). As shown in Fig. \ref{fig7}, PVTC consistently outperforms alternative methods across all compression ratios. At 256 tokens, our method enhances the performance over Pixel-shuffle, LDPv2, and TokenPacker by 2.25\%, 1.71\%, and 1.01\%, respectively, and achieves results comparable to the linear projector, reaching the performance limit without token compression. We attribute this to PVTC's dual-query design, which allows the compressed visual tokens to effectively capture global semantics and local details.




To further understand the individual contributions of each component in PVTC, we perform an ablation study focusing on local and global queries, particularly examining the effect of different attention layers $L$ on the results. When both layers are set to 0, this configuration corresponds to the baseline setup of InternVL2. As indicated in Table \ref{table3}, utilizing either query type alone enhances performance compared to the baseline model, while combining both queries achieves the best performance. When using only the CLS token for querying, the model experiences limited improvements, while introducing learnable CLS scales significantly boosts the performance of global query. These results demonstrate that local and global queries offer complementary advantages, each capturing distinct aspects of visual information.
We find that PVTC performs optimally with 1$\sim$2 layers. The performance tends to decline when $L > 2$. To balance the parameters and performance, we ultimately set $L = 1$.

\begin{table}[t]
    \vspace{1.2ex}
    \centering
    \setlength{\abovecaptionskip}{6pt}
    \setlength{\belowcaptionskip}{-14pt}
    \resizebox{\linewidth}{!}
    {
        \begin{tabular}{@{}l@{\hskip 0.4em}@{\hskip 0.4em}l@{\hskip 0.4em}@{\hskip 0.4em}l@{\hskip 0.4em}@{\hskip 0.4em}l@{\hskip 0.4em}@{\hskip 0.4em}l@{\hskip 0.4em}@{\hskip 0.4em}l@{\hskip 0.4em}@{\hskip 0.4em}l@{\hskip 0.4em}@{\hskip 0.4em}l@{\hskip 0.4em}@{\hskip 0.4em}l@{\hskip 0.4em}@{\hskip 0.4em}l@{}}
        \toprule
        \multirow{2}{*}[0pt]{\makecell[c]{Method}}  &
                                                    Global      &Local       &Text    &Doc     &GQA     &Viz    &MM     &PO     &\multirow{2}{*}[0pt]{\makecell[c]{Avg}} \\
                                                &    Attn        &Attn        &VQA     &VQA      &       &Wiz    &Vet    &PE \\
        \midrule
                                       Base         &    0           &0           &44.99   &23.21   &61.03   &46.55  &24.6   &86.33  &47.78   \\ 
        \midrule
        \multirow{6}{*}[0pt]{\makecell[c]{PVTC}}  &   
                                                    1           &0           &44.52   &22.57   &60.52   &47.13  &23.1   &85.31  &47.19   \\
                                                 &    1*          &0           &45.33   &22.98   &61.35   &48.79  &25.4   &86.21  &48.34   \\
                                                 &    0           &1           &45.76   &23.44   &61.64   &51.33  &26.8   &86.98  &49.32   \\
                                                    &    1*          &1           &47.69   &24.82   &61.81   &51.41  &27.8   &86.67  &50.03   \\
                                                    &    2*          &2           &47.92   &25.11   &61.63   &52.33  &27.6   &87.21  &50.30   \\
                                                    &     3*          &3           &47.51   &24.76   &60.97   &49.88  &27.9   &87.03  &49.67   \\
        \bottomrule
        \end{tabular}
    }
    \caption{Ablation experiments on global attention and local attention. * indicates that use learnable CLS scale.}
    \label{table3}
\end{table}

\textbf{LVTC.}
We set $M_{1}=2$ (corresponding to 256 tokens), $M_{2}=1$ (corresponding to 1024 tokens) and $k$ as the 3/4 layer of LLM, where $k=17$ for 2B model and $k=23$ for 8B model. We adopt the Pixel-shuffle + MLP as basic projector.  
For base model, we conduct experiments under 3 distinct compression settings: LR compression (256), HR compression (1024), and HR$\rightarrow$LR compression (1024$\rightarrow$256). 
For LVTC, we initially apply LR compression and expand it to HR compression at layer $k$. Additionally, we assess the impact of the HR projector and multi-projector components. 
The experimental findings, summarized in Table \ref{table4}, lead to three conclusions:
1. The model’s performance is mainly influenced by the token numbers, with more tokens generally yielding better results.
2. Backward compression is more effective than forward compression, indicating that visual tokens primarily contribute in deeper layers. 
3. Upsampling to increase tokens doesn't enhance performance, while integrating information from HR projector and multi-projector significantly boosts it. 
In the 2B model, these two modules individually contribute performance improvements of 1.87\% and 0.98\%, respectively, while their combined effect brings a total gain of 2.32\%, excedding the performance without token compression.

\begin{table}[h]
    \setlength{\abovecaptionskip}{6pt}
    \setlength{\belowcaptionskip}{-10pt}
    \centering
    \resizebox{0.9\linewidth}{!}
    {
        \begin{tabular}{@{}l@{\hskip 0.5em}@{\hskip 0.5em}l@{\hskip 0.5em}@{\hskip 0.5em}l@{\hskip 0.5em}@{\hskip 0.5em}l@{\hskip 0.5em}@{\hskip 0.5em}l@{\hskip 0.5em}@{\hskip 0.5em}l@{\hskip 0.5em}@{\hskip 0.5em}l@{}}
        
        \toprule
        \multirow{2}{*}[0pt]{\makecell[c]{Model}} &
        \multirow{2}{*}[0pt]{\makecell[c]{Method}}             &Token            &Token               &HR         &Multi        &\multirow{2}{*}[0pt]{\makecell[c]{Accuracy}} \\
                                        &                      &$0 \sim k-1$       &$k \sim last$    &proj           &proj                                                  \\
        \midrule
        \multirow{7}{*}[0pt]{\makecell[c]{InternVL2-2B}}        &
        \multirow{2}{*}[0pt]{\makecell[c]{Base}}               &256              &256            &                  &              &47.78  \\
        &                                                      &1024             &1024           &                  &              &49.93  \\
        &                                                      &1024             &256            &                  &              &48.05  \\
        \cmidrule(lr){2-7}
         & \multirow{4}{*}[0pt]{\makecell[c]{LVTC}}        
                                                               &256              &1024           &                   &             &47.91  \\
                &                                              &256              &1024           &\checkmark         &             &49.78  \\
                &                                              &256              &1024           &                   &\checkmark   &48.89  \\
                &                                              &256              &1024           &\checkmark         &\checkmark   &50.23  \\
        \midrule
        \multirow{4}{*}[0pt]{\makecell[c]{InternVL2-8B}}     &
        \multirow{2}{*}[0pt]{\makecell[c]{Base}}               &256              &256            &                  &              &51.85  \\
        &                                                      &1024             &1024           &                  &              &53.56  \\
        \cmidrule(lr){2-7}
         & \multirow{2}{*}[0pt]{\makecell[c]{LVTC}}   
                                                               &256              &1024           &\checkmark         &             &53.38  \\
                &                                              &256              &1024           &\checkmark         &\checkmark   &53.77  \\
        \bottomrule
        \end{tabular}
    }
    \caption{Component-wise ablation study of LVTC.}
    \label{table4}
\end{table}

\begin{figure}[h]
    \setlength{\abovecaptionskip}{4pt}
    \setlength{\belowcaptionskip}{-10pt}
    \centering
    \includegraphics[width=3.25in]{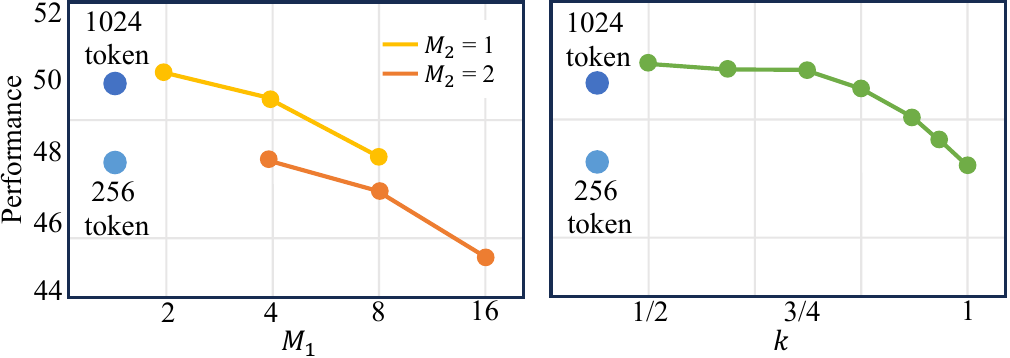}
    \caption{Ablation experiments on, left: LR compression and HR compression multiples, right: token expansion layer.} 
    \label{fig8}
\end{figure}

Futhermore, we explore the impact of parameters $M_{1}$, $M_{2}$, and $k$. As illustrated in Fig. \ref{fig8}, the performance upper limit and computational bottleneck of the model are primarily influenced by $M_{2}$. Increasing $M_{1}$ will lead to performance degradation. 
With setting $M_{1}=2$, $M_{2}=1$, as $k$ decreases from 1 to 3/4, the model’s performance demonstrates an almost linear improvement. Afterwards, the metrics slightly increase and then stabilize. When $k$ reaches 1/2, the performance becomes essentially saturated.

\textbf{RVTC.}
We utilize various image slicing methods for verification, with a dataset of 770K training samples and a maximum patch size set to 6. We compare the average number of patches, average token, training time, and performance across ratio matching, area matching, and edge length matching methods, as shown in Table \ref{table5}.
Compared to the baseline method, the area matching and edge length matching strategies on the 2B model require only 49.62\% and 58.51\% of the training time, respectively, while achieving 97.77\% and 99.04\% of the performance of the ratio matching method.  
The conclusions on the 8B model are similar, with just 50.69\% and 59.88\% of the training time, our methods could reach 97.98\% and 99.26\% of the baseline model’s performance, respectively. By maximizing the utilization of visual information, our proposed method significantly enhances model training efficiency with slight performance loss.



\begin{table}[h]
    \vspace{1ex}
    \centering
    \setlength{\abovecaptionskip}{6pt}
    \setlength{\belowcaptionskip}{-10pt}
    \resizebox{0.9\linewidth}{!}
    {
        \begin{tabular}{@{}l@{\hskip 0.5em}@{\hskip 0.5em}l@{\hskip 0.5em}@{\hskip 0.5em}l@{\hskip 0.5em}@{\hskip 0.5em}l@{\hskip 0.5em}@{\hskip 0.5em}l@{\hskip 0.5em}@{\hskip 0.5em}l@{}}
        
        \toprule
        \multirow{2}{*}[0pt]{\makecell[c]{Model}} &
        \multirow{2}{*}[0pt]{\makecell[c]{Method}}                  &Average       &Average            &Train              &\multirow{2}{*}[0pt]{\makecell[c]{Accuracy}} \\
                                        &                           &patch         &token           &time               &                                             \\
        \midrule
        \multirow{3}{*}[0pt]{\makecell[c]{InternVL2-2B}}
                &Base-ratio                                         &5.009         &1282              &13.5h              &58.77       \\
                &RVTC-area                                          &1.559         &399           &6.7h               &57.46   \\
                &RVTC-edge                                          &2.033         &520           &7.9h               &58.21 \\
        \midrule
        \multirow{3}{*}[0pt]{\makecell[c]{InternVL2-8B}}
                &Base-ratio                                         &5.009         &1282             &35.9h             &63.45      \\
                &RVTC-area                                          &1.559         &399           &18.2h             &62.17  \\
                &RVTC-edge                                          &2.033         &520           &21.5h             &62.98    \\
        \bottomrule
        \end{tabular}
    }
    \caption{Experimental comparison of different slicing methods.}
    \label{table5}
\end{table}

\section{Conclusion}
\vspace{-1ex}



This paper presents the InternVL-X series, showcasing three innovative visual token compression modules: PVTC, LVTC, and RVTC. These modules significantly enhance model effectiveness and training optimization. PVTC employs a grid-based cross-attention approach to effectively balance local detail with global semantics. LVTC compresses tokens early and expand them in deep layers to enhance LLM efficiency. Meanwhile, RVTC optimizes high resolution image slicing based on pixel counts or edge length, thereby boosting training speed with minimal impact on performance. 
Overall, InternVL-X models achieve state-of-the-art results across multiple benchmarks, outperforming other open-source models and delivering substantial improvements in both performance and efficiency.

{
    \small
    \bibliographystyle{ieeenat_fullname}
    \bibliography{main}
}

\clearpage
\onecolumn

\begin{appendices}
    \section{Visual results}
    \vspace{-2ex}

    \begin{figure*}[h]
        \setlength{\abovecaptionskip}{2pt}
        \setlength{\belowcaptionskip}{-10pt}
        \centering
        \includegraphics[width=6.4in]{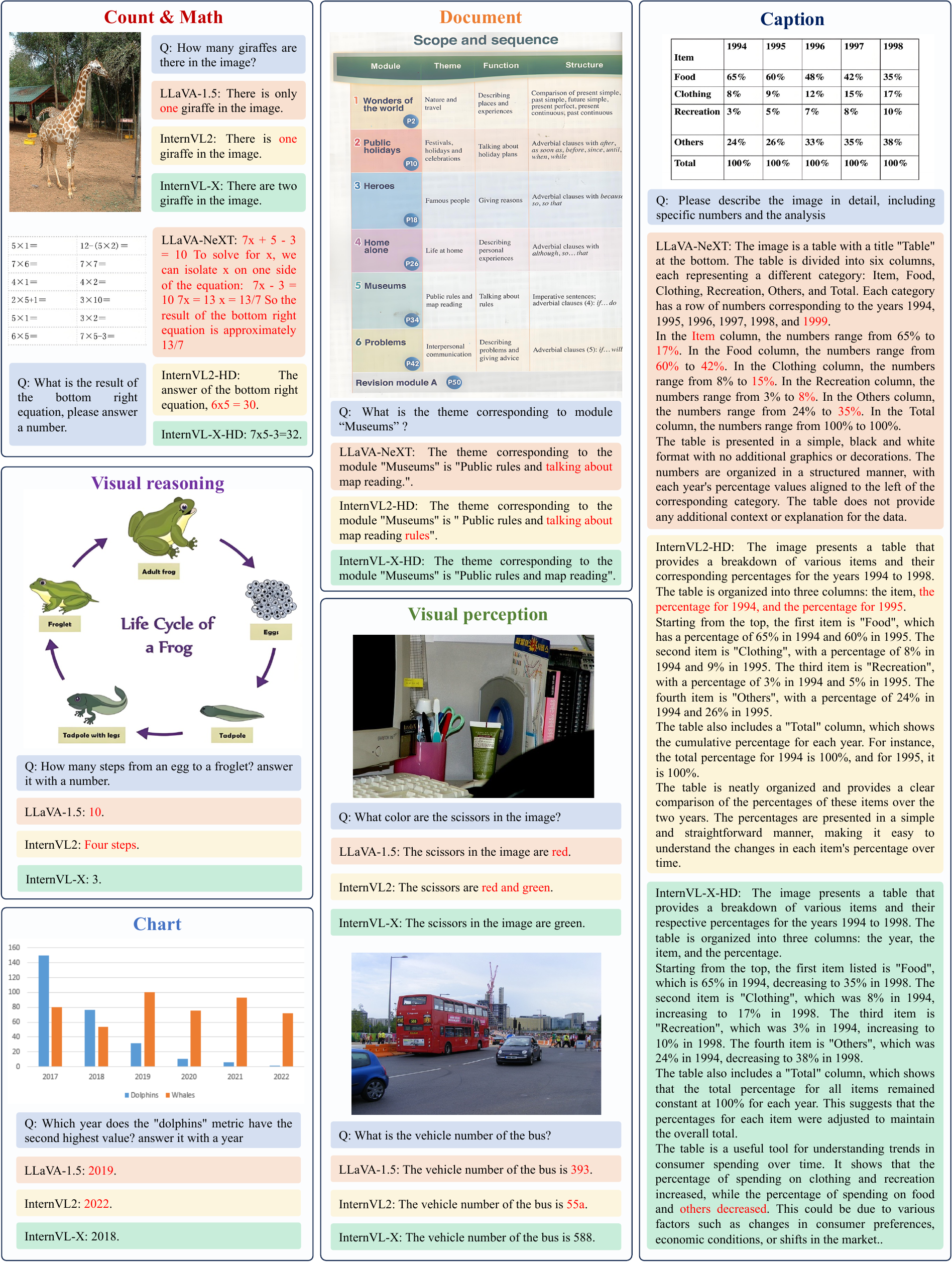}
            \caption{Visual comparison of InternVL-X with other models. Red fonts represent incorrect answers.}
            \label{fig10}
    \end{figure*}
\end{appendices}

\end{document}